\documentclass[review,times,onecolumn,authoryear]{elsarticle}
\usepackage{lineno,hyperref}
\usepackage{amsmath}
\usepackage{float}
\usepackage{apalike}
\usepackage{xcolor}
\usepackage{multicol}
\usepackage{amssymb}
\usepackage{latexsym}
\usepackage{algorithm}
\usepackage{framed,multirow}
\usepackage{amssymb}
\usepackage{latexsym}

\usepackage{algorithm}
\usepackage{algorithmic}
\modulolinenumbers[1]

\journal{to Expert Systems with Applications}

\biboptions{authoryear}

\begin{document}

\begin{frontmatter}

\title{A Novel Capsule Neural Network Based Model for Drowsiness Detection Using Electroencephalography Signals}

\author{Luis Guarda\corref{cor1} \fnref{fn1}}
\author{Juan Tapia\corref{cor1} \fnref{fn2}}\ead{juan.tapia-farias@h-da.de} 
\author{Enrique Lopez Droguett\corref{cor1} \fnref{fn1}}\ead{elopezdroguett@ing.uchile.cl }
\author{Marcelo Ramos\corref{cor3} \fnref{fn3}}\ead{mrmartin@usp.br} 
\cortext[fn2]{Corresponding author. E-mail address: juan.tapia-farias@h-da.de}
\address[fn1]{Mechanical Engineering Department, University of Chile, Santiago, Chile}
\address[fn2]{da/sec-Biometrics and Internet Security Research Group, Hochschule Darmstadt, Germany.}
\address[fn3]{Analysis, Evaluation and Risk Management Laboratory – LabRisk, Naval Architecture, Sao Paulo, Brazil.\\
**\textcolor{blue}{This is a pre-print version. Full version will be available in Expert Systems With Application**}
}

\begin{abstract}
The early detection of drowsiness has become vital to ensure the correct and safe development of several industries' tasks. Due to the transient mental state of a human subject between alertness and drowsiness, automated drowsiness detection is a complex problem to tackle. The electroencephalography signals allow us to record variations in an individual's brain's electrical potential, where each of them gives specific information about a subject's mental state. However, due to this type of signal's nature, its acquisition, in general, is complex, so it is hard to have a large volume of data to apply techniques of Deep Learning for processing and classification optimally. Nevertheless, Capsule Neural Networks are a brand-new Deep Learning algorithm proposed for work with reduced amounts of data. It is a robust algorithm to handle the data's hierarchical relationships, which is an essential characteristic for
work with biomedical signals. Therefore, this paper presents a Deep Learning-based method for drowsiness detection with CapsNet by using a concatenation of spectrogram images of the electroencephalography signals channels. The proposed CapsNet model is compared with a Convolutional Neural Network, which is outperformed by the proposed model, which obtains an average accuracy of 86,44 \% and 87,57\% of sensitivity against an average accuracy of 75,86\% and 79,47\% sensitivity for the CNN, showing that CapsNet is more suitable for this kind of datasets and tasks.

\end{abstract}

\end{frontmatter}

\nolinenumbers



\section{Introduction}
\label{sec:introduction}

In the last decades there has been a sustained increase in the level of automation of work environments, which has meant that it becomes increasingly common the execution of monotonous tasks that require a high level of attention \citep{hockey2003operator}. When these tasks are intended to ensure the correct and safe operation in work environments such as control rooms, it is essential to ensure that operators maintain an adequate level of alertness. In this context, the early detection of drowsiness in operators has become of vital importance to guarantee the reliability of the operation.

Drowsiness is defined as a middle state between slumber and alertness, where the subject has an overwhelming desire to sleep. These symptoms are accompanied by an increase in the reaction times of the subject and, consequently, a reduction in his or her surveillance capacity. As an example, in the incidents of Three Mile Island and Chernobyl, drowsiness was documented as a contributing factor.  In this context, drowsiness is usually used interchangeably with the term mental fatigue, even though mental fatigue is considered a broader concept \citep{Leonard_Trejo, Ramzam}.

Through the use of electroencephalography signals (captured from several electrodes located on the scalp), it is possible to record the bioelectrical activity generated by the cortical neurons of the brain and their variation in time, which provide information about the mental state of the subject. Particularly, the measurement of the frontals and parietals channels can be used to analyze the level of drowsiness of a subject due the variations on the amplitude in the frequencies range of theta and alpha waves that can be tracked in these channels \citep{Leonard_Trejo, BARUA2019121}. On the other hand, studies have found that there exist a relationship between the amplitude variations of alpha and theta waves in a drowsy subject with the Karolinska sleepiness scale \citep{KAIDA20061574}, which measures the level of self-perceived sleepiness of an individual into nine scores \citep{Akerstedt}.
However, the detection of drowsiness has not been limited to the use of electroencephalography signals. Several authors have used parallel and / or complementary inputs such as the use of electrocardiography signals \citep{Chui}, and eye tracking devices \citep{YEO2009115}, through the use of machine learning algorithms. The main limitation of this type of methods is that they are based on the extraction and selection of features in a handcrafted way, which are subject to the uncertainty and bias of the experts in charge of the manual feature engineering. Although the methods of eye tracking have achieved good results, the use of video devices makes these techniques too invasive for the operator’s privacy.

In this context, Deep Learning techniques are presented as a valid alternative to replace the methods of extraction and selection of handcrafted features, thanks to their ability to learn how to identify specific features of their environment autonomously \citep{LeCun}, thus avoiding the need to have experts in each area that direct the learning.

Using Deep Learning algorithms, several authors have used EEG signals to achieve the detection of different classification tasks. As an example, \citep{Zheng} use a Deep Neural Network in combination with a six-electrode EEG to classify human emotions, and in \citep{Spampinato} authors use Recurrent Neural Networks (RNN) and a Convolutional Neural Network (CNN) to learn visual stimuli-evoked data from EEG signals and classify and then reconstruct object classes from the ImageNet dataset \citep{Deng}. 

Nonetheless, the use of Deep Learning algorithm has not been limited to the use of raw signals as input. Spectrograms of the corresponding signals as input data have attracted many authors \citep{Yuan, Lees2019DeepLC,NIPS2009_3674}. In this context, Verstraete et al. developed a CNN based model \citep{Verstraete2017DeepLE} able to carry out the fault diagnosis of ball bearings through the use of spectrograms from vibration signals of the rolling elements analyzed, achieving better results in the classification task than conventional shallow methods of manual feature extraction and selection. 

However, Convolutional Neural Networks have shown to have two main problems. First, they require large volumes of data for training, which is an inherent problem when working with bioelectric signals. Secondly, due to the loss of information that occurs in the pooling layer of the CNN's, there is a positional invariance of the components that the network uses to carry out the classification, which prevents it from being sensitive to changes of rotation or translation within the image \citep{NIPS2017_6975}. One way to eliminate this problem is with excessive training for all of the possible angles, but it usually takes a greater dataset and a lot more time and computational resources. This positional invariance represents a problem when it comes to dealing with transient 1-D signals such as electroencephalography signals.

In this context, Capsule Neural Networks (CapsNet) were recently introduced as a Deep Learning technique which seeks to solve CNNs problems through the replacement of the pooling layer by a routing-by-agreement algorithm and the replacement of the use of scalars within the network by capsules (vectors) \citep{NIPS2017_6975}. These changes allow this type of network to work with a smaller amount of data and can establish equivalence relationships between the characteristics of the images.

The CapsNet architecture introduced in \citep{NIPS2017_6975} not only allowed us to solve the inherent problems of CNNs, but it also reached state-of-the-art results for the MNIST dataset\citep{mnist}. For this reason, CapsNet based models have seen a surge in the recent literature with applications in a variety of fields as a variety of areas such as the work of Xi et al \citep{Xi,e2018matrix}.

Moreover, in \citep{Upadhyay2018GenerativeAN}, the authors proposed a Generative Adversarial Networks (GANs) that employs a CapsNet as the discriminator. Rawlinson et al. \citep{Rawlinson} perform an unsupervised training of latent capsule layers using only the reconstruction loss as input, obtaining a better generalization than the supervised model for the affNIST and MNIST benchmarks datasets. CapsNets have been also applied to process biomedical data as, for example, in \citep{LaLonde2018CapsulesFO} where the authors expand CapsNets to work with large images of lungs computed tomography scans, and Afshar et al. \citep{Afshar} use CapsNets to tackle the brain tumor type classification task, outperforming the performance of CNN in the same problem.

Due the aforementioned limitations of CNNs to work with reduced volumes of data and for being incapable of handling changes in rotation or translation of the input data, we propose a Capsule Neural Network that is able to work with EEG signals in order to tackle the problem of drowsiness detection, thus using the advantages of CapsNet over CNNs to achieve a higher capability of generalization and accuracy, and also avoiding the manual extraction of features which introduce a bias to the algorithm.

The proposed Capsule Neural Network model for drowsiness detection makes use of spectrograms from electroencephalogram signals as input. Moreover, the spectrograms were obtained from the frontal Fz and parietal Pz channels due to the relationship between these channels and the variations in amplitude of alpha and theta waves when a subject is under a drowsiness state \citep{Leonard_Trejo}. Also, two spectrograms datasets are considered: the first one corresponds to spectrograms from EEG signals of the Fz channel, whereas the second dataset corresponds to the vertical concatenation of spectrograms obtained from both the Fz and Pz channels.

The proposed CapsNet model’s performance in the drowsiness detection task is also compared to a deep Convolutional Neural Network model as well as other shallow machine learning models such as Neural Networks \citep{SCHMIDHUBER201585}, Support Vector Machines \citep{Cortes1995} and Random Forests \citep{Breiman}.

The paper is organized as follows. Section 2 provides an overview about the background of drowsiness, electroencephalography, CNNs and CapsNets. Section 3 presents the datasets used to carry out the detection of drowsiness and the pre-processing methodology applied to the EEG signals as well as comparisons of the proposed CapsNet model with CNN, SVM, Random Forest and shallow neural network. Section 4 present the experiments. Section 5 the results and discussion, finally section 6 with the concluding remarks.

\section{Background}
\subsection{Drownsiness}
Drowsiness can be described as the set of psychological and physiological changes that a subject suffers while executing a task that requires a high level of attention for a great period of time. Drowsiness is manifested as a deterioration in the cognitive and psychomotor proficiency of the individual, which results in a reduced alertness, accompanied by a lack of energy \citep{Tadeusz}. In this way, drowsiness can be understood as the loss of a subject's basic abilities, such as reacting in a timely and appropriate manner to an unforeseen event \citep{ASCH}.

When an individual is in a state of drowsiness, the following symptoms may appear:

\begin{itemize}
    \item Slow or clumsy movements.
	\item Decrease in the motor speed of reaction.
	\item Appearance of blurred or double vision.
	\item Difficulty remembering.
	\item Difficulty concentrating or staying alert.
\end{itemize}

From this perspective, an individual who works under the effects of drowsiness in an environment that requires a high level of attention can provoke an accident \citep{ASCH}.

\subsection{Electroencephalography}

Electroencephalography (EEG) is an electrophysiological monitoring method that allows the recording of the bioelectrical activity generated by the cortical neurons of the brain and their variation over time. This technique is carried out by placing small electrodes on the subject's scalp. The signals picked up by the electrodes are delivered to an electroencephalograph, which amplifies the brain wave potentials \footnote{\url{https://www.trans-cranial.com/docs/10_20_pos_man_v1_0_pdf.pdf}} 

To guarantee the standardized reproducibility of the captured EEG signals, the international 10-20 electrode placement system\citep{Homan}, which is based on the relationship between the location of an electrode and the underlying area of the electrode cerebral cortex. Figure 1 shows the traditional disposition of the electrodes for the international system 10-20.

The EEG signals can be divided into four segments depending on their frequency band. Each of these bands refers to different states of consciousness such as sleep, intense mental activity, and drowsiness. These four frequency bands are the following:

\begin{itemize}
    \item Delta (0.5-4Hz): its appearance occurs mainly during deep sleep. 
    \item	Theta (4-8Hz): this type of wave is characterized by a state of drowsiness of the individual with reduced consciousness. 
    \item	Alpha (8-13Hz): it represents a state of low brain activity and relaxation both physical and mental, but aware of the environment. 
    \item	Beta (13-20Hz): waves of this type are emitted when a subject is in a conscious or alert state. Beta waves denote intense mental activity.
    \item   Gamma ($>$ 30Hz): its appearance is associated with large-scale brain network activity such as attention and working memory\citep{Nie}.
\end{itemize}
In the context of drowsiness detection, the frontal and parietal channels provide information about the level of drowsiness of an individual, which is represented in variations of the alpha and theta waves \citep{Leonard_Trejo}.

\begin{figure}[h!]
\centering
\includegraphics[scale=0.8]{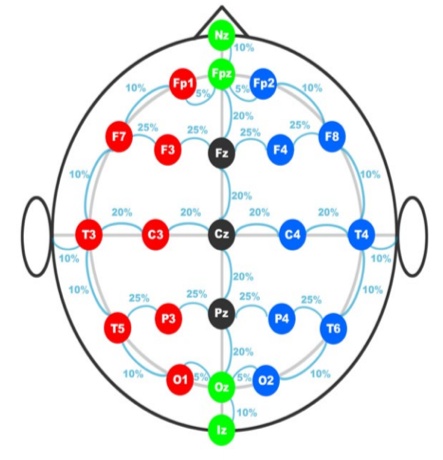}
\caption{International electrode placement system 10-20 \citep{Homan}.}
\label{fig:universe}
\end{figure}

\subsection{Karolinska Sleepiness Scale}
In order to avoid detection of fatigue events where the subject perceives himself as alert (False positive), it is possible to label the EEG data for sleepiness using the Karolinska Sleepiness Scale (KSS). This scale measures the level of self-perceived sleepiness for the individual in the last 10 minutes, and a significant relationship has been found between the variation of alpha and theta waves and the KSS \citep{KAIDA20061574,Akerstedt}. Table \ref{table1} shows the description of each level of the scale, which has a range from 1 to 9.

\begin{table}[H]
\centering
\caption{Karolinska Sleepiness Scale {[}4{]}.}
\label{table1}
\begin{tabular}{|c|c|}
\hline
\textbf{Scale} & \textbf{Psychophysical state}            \\ \hline
1.             & Extremely alert                          \\ \hline
2.             & Very alert                               \\ \hline
3.             & Alert                                    \\ \hline
4.             & Rather alert                             \\ \hline
5.             & Neither alert nor sleepy                 \\ \hline
6.             & Some sighs of sleepiness                 \\ \hline
7.             & Sleepy, but no difficult remaining awake \\ \hline
8.             & Sleepy, some effort to keep alert        \\ \hline
9.             & Extremely sleepy, fighting sleep         \\ \hline
\end{tabular}
\end{table}

\subsection{Convolutional Neural Network}
Convolutional Neural Networks (CNN) are a Deep Learning algorithm developed to tackle matrix (structures) data such as images. CNNs are composed of three types of layers: convolution layer, pooling layer and classification layer \citep{Goodfellow}. Due to this arrangement, CNNs can reduce dimensionality and learn specific non-linear representations. The specific function of each layer can be summarized as follows (see Figure \ref{figure2} and Figure \ref{figure3}):

\begin{itemize}
\item{\textbf{Convolutional layer}}: in this first layer, a convolution between the input image and multiple local filters is made, generating different invariant local features for each kernel (filter) applied. This process allows certain characteristics of the input data to become more dominant in the output. In Equation (1), a convolution between a two-dimensional input (I) and a two-dimensional kernel (K) is shown:

\begin{equation}
    S(i,j)=(I*K)(i,j)=\sum_m \sum_n I(i+m,j+n)K(m,n)
\end{equation}

\item{\textbf{Pooling layer:}}
usually the pooling layers are located after a convolutional layer. The function of this type of layer is to extract the most significant features of the output of the previous layer. This features extraction is done performing a sampling reduction, which implies a loss of information\citep{NIPS2017_6975}, but it also leads to a lower calculation overload, and introduces translational invariance to the model, which allows the model to be less susceptible to small changes in the input, reducing the possibility of overfitting \citep{Boureau}. The sampling reduction that occurs in the pooling layers implies that the network is not able to reconstruct the input after a pooling layer because the characteristics that were not selected are dropped and cannot be used again. In the same way, after a pooling layer, the network forgets wherein the data and how many times (inside the pooling filter size) each feature appears. On the other hand, the max-pooling function is one of the possible choices to use in this layer. This operation finds the maximum value between sample windows, saving only these values for the next layer. In Figure \ref{figure2}, the operation performed by the max-pooling can be observed graphically.

\begin{figure}[h!]
\centering
\includegraphics[scale=0.8]{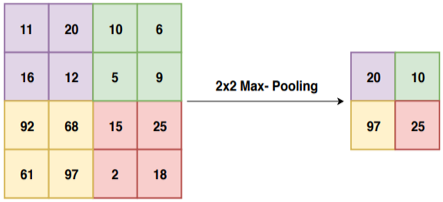}
\caption{Example of 2x2 max pooling.}
\label{figure2}
\end{figure}

\item{\textbf{Classification layer}} subsequently the convolution and pooling layers, one or more fully connected layers of neurons can be used, in which each pixel of the last output layer is considered as an input to an individual neuron. Using this input, the fully connected layer performs the desired classification.

\begin{figure*}[h!]
\centering
\includegraphics[scale=0.8]{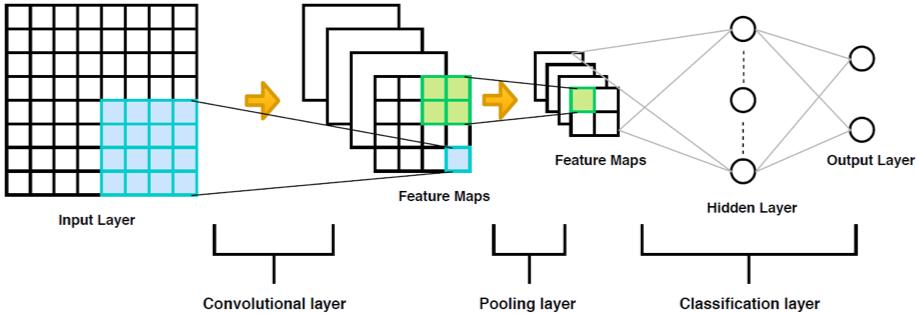}
\caption{Generic CNN Arquitecture.}
\label{figure3}
\end{figure*}
\end{itemize}

\subsection{Capsule Neural Network: An Overview}

Through the pooling operation, CNNs are able to perform dimensionality reduction and to add a certain degree of translational invariance to the model. This results in a reduction of the required computational cost in classification (i.e., diagnostics) tasks, whereas the translational invariance allows the model to identify small changes in the inputs, improving its generalization capability \citep{Boureau}.

However, the above-mentioned aspects of CNNs have some inherent problems. Indeed, the translational invariance only makes the network less sensitive to small changes of translation, but it is not able to deal with rotations or different angles of the same input. Also, due to the loss of information that takes place in the dimensionality reduction operation, the network cannot keep the spatial and hierarchical relationships among the extracted features.

\begin{figure*}[h!]
\centering
\includegraphics[scale=0.8]{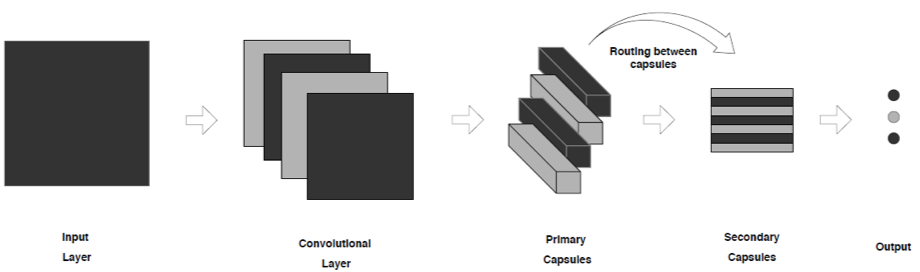}
\caption{Generic Capset architecture.}
\label{figure4}
\end{figure*}

In order to tackle these inherent limitations of CNNs, Sabour et al \citep{NIPS2017_6975} introduced a novel algorithm for image recognition known as Capsule Neural Networks, which incorporates the concept of equivariance to the network, allowing the model to keep the spatial and hierarchical relationships among the extracted features. The generic architecture of CapsNet is composed by the following layers: input layer, convolutional layer, primary capsules, secondary capsules and finally the output as shown in Figure \ref{figure4}.

In this model, capsules are groups of neurons represented as vectors, where each neuron denotes a feature from an object and the length of the vector corresponds to the capsule’s activation probability. To ensure that the vector length is between 0 and 1, a squashing function is applied as follows:

here $v_j$ is the “squashed” value of the capsule’s output $s_j$.
\begin{equation}
    v_j=\frac{‖s_j ‖^2}{(1+‖s_j ‖^2 )}\times \frac {s_j}{‖s_j ‖^2} 
\end{equation}

To get the first layers of capsules, the algorithm performs a convolution operation, with the extracted features maps being rearranged into vectors (with a previously given dimension) and are resized using Equation (2).

Then, in order to select which features are more predominant for the current prediction and thus avoiding the pooling process, Capsule Neural Networks use an iterative process called dynamic routing, which takes the predictions from  primary capsules and computes the actual output according to an agreement metric between predictions. 

The predictions of the actual output for each capsule are calculated using Equation (3):

\begin{equation}
\hat{u}_{(j|i)}=w_{ij} u_i
\end{equation}

where $u_i$ is the output of capsule $i$, $\hat{u}_{(j|i)}$ is the prediction of the $j^t$ higher capsule given $u_i$ and $W_{ij}$ is the weighting matrix between capsule $i$ and capsule $j$. The output of the $j\hat{th}$ higher capsule $s_j$ corresponds to a weighted sum over all $\hat{u}_(j|i)$, which is obtained using Equation (4) as:

\begin{equation}
    s_j=\sum_i c_{ij} \hat{u}_{(j|i)}
\end{equation}

where $c_{ij}$ are called coupling coefficients, and are calculated by Equation (5):

\begin{equation}
  c_{ij}=\frac{exp(b_{ij})}{\sum_k exp(b_{ik})}
\end{equation}

On the other hand, the coupling coefficients are calculated in an iteratively way by the dynamic routing process using Equation (6):

\begin{equation}
b_{ij}  \leftarrow  b_{ij}+\hat{u}_{(j|i)}*v_j
\end{equation}

where $\hat{u}_{(j|i)}*v_j$ indicates the level of agreement between the actual output of the capsule and the prediction done by the corresponding capsule in the first layer \citep{NIPS2017_6975}.

The number of iterations is previously defined, according to the classification task that is desired. Finally, the classification is made according to the prediction made by the capsules in the last iteration.

\section{The Dataset}

In this study, only the EEG signals and the KSS scores of the subjects were used. From the five EEG channels available for each subject, only the signals recorded by the frontal (Fz) and parietal (Pz) electrodes were used. This choice is based on the fact that alpha and theta waves show an increase of amplitude that can be measured in mentioned channels in case of the subjects is in the  drowsy \citep{Leonard_Trejo}. 

In order to generate the labels of the EEG signals, the description of the KSS indexes was used. In this way, the data of the subjects that reported a KSS index (see Table 1) less than or equal to 4 were characterized as alert (no drowsiness), and the subjects who reported a KSS index greater than or equal to 7 were characterized as drowsy \citep{Akerstedt}. 
This labeling rule was applied only for the data from PVT 1 and PVT 3 (see Figure \ref{figure7}), excluding the data from PVT 2 due to the average values of the KSS index of the PVT 2 $(5,6+-1,7)$, which do not correspond to any of the classes previously defined. On the other hand, it should be noted that none of the KSS indices of the PVT 3 corresponds to the alert state and none of the KSS indices of PVT 1 corresponds to the drowsy state. Thus, all the signals labeled as alert correspond to the PVT 1 and all the signals labeled as drowsy were obtained from the PVT 3. Once this is done, the data from 10 subjects from PVT 1 and the data from 10 subjects from PVT 3 satisfy the mentioned labeling rule.
After the labeling process and with the aim of increasing the data size, the EEG signals (which originally had a duration of 10 minutes) from both channels Fz and Pz were segmented into files of thirteen seconds durations. This time interval corresponds to the minimal duration that an EEG signal must have to enclose all the subject mental alertness \citep{Leonard_Trejo}. After this, the amount of data from each channel was 920 files with 460 for each class.

Then, a spectrogram was obtained for each file of thirteen seconds. The spectrograms were generated using 512 points as window size, an overlap of 50\% and plotting only the frequencies from 0 to 20 Hz, given that one is interested only in the variations of alpha and theta waves \citep{Leonard_Trejo}. 
The spectrograms were converted into grayscale, resulting in a size of $32\times32\times 1$ pixels and the color bar and axes were removed.

Based on the abovementioned procedure, two sets of spectrograms were obtained, which are described below:
\begin{enumerate}
    \item Spectrograms from channel Fz / 13 seconds duration: set of spectrograms from channel Fz with size of 32x32 pixels, a duration of thirteen seconds and a frequency range from 0 to 20 Hz.
    \item Spectrograms from channels Fz-Pz / 13 seconds duration: set of spectrograms from both channels Fz and Pz concatenated vertically, resulting in $64\times32$ pixels, with a duration of thirteen seconds and a frequency range from 0 to 20 Hz.
\end{enumerate}

\begin{figure}[h!]
\centering
\includegraphics[scale=0.35]{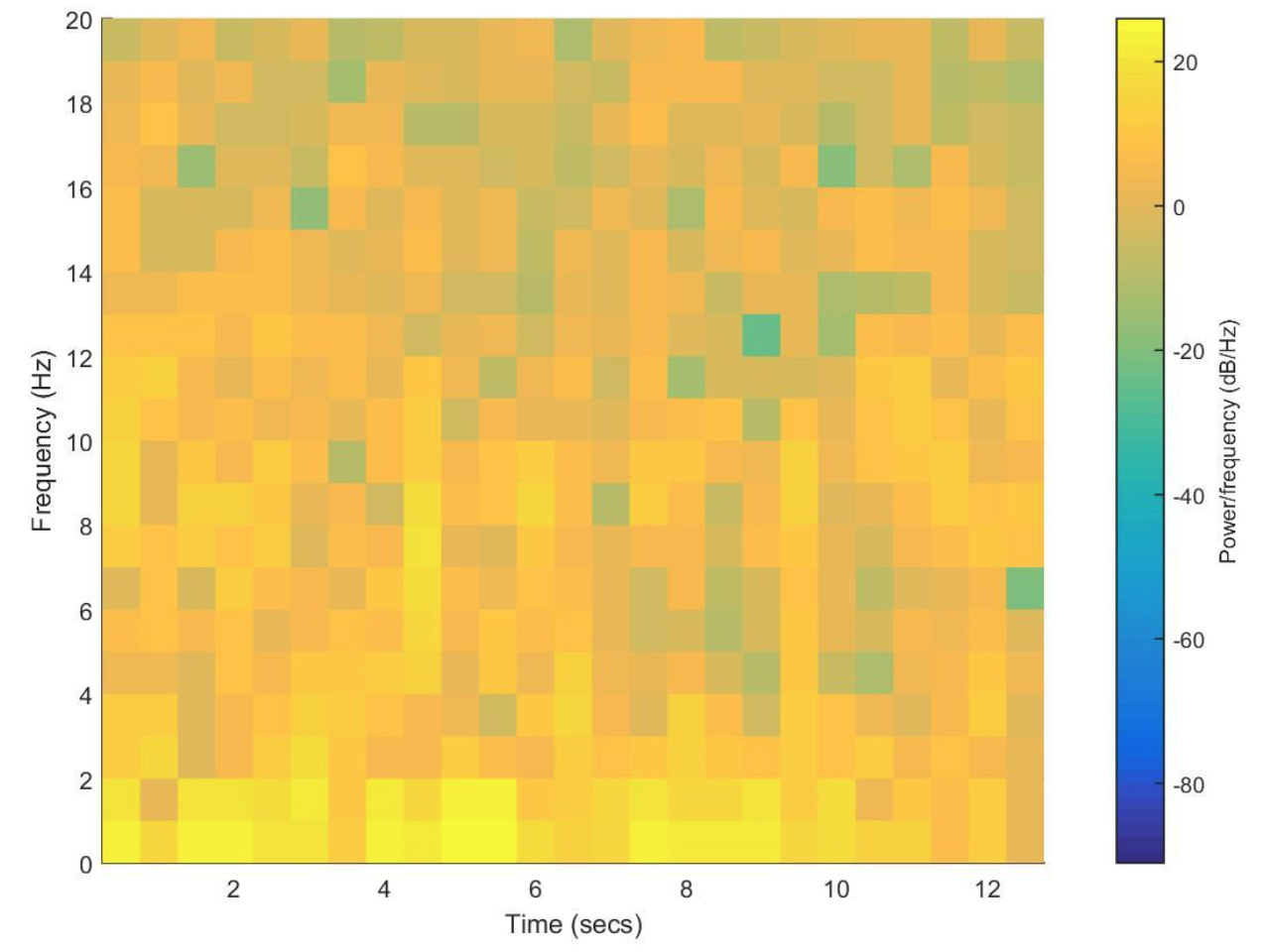}
\caption{Example of EEG spectrogram obtained using a windows size of 512 points, an overlap of 50\% and 512 discrete Fourier transforms points (NFFT).}
\label{figure8}
\end{figure}

Figure \ref{figure8} show a spectrogram using a 512 windows size. A spectrogram was obtained for each file of thirteen seconds using the raw signal. The spectrograms were generated using 512 points as window size, an overlap of 50\%, 512 discrete Fourier transform points (NFFT) and plotting only the frequencies from 0 to 20 Hz, given that we are interested only in the variations of alpha and theta waves. 
The spectrograms were converted into grayscale, resulting in a size of $32 \times 32 \times 1$ pixels and the color bar and axes were removed.
 
\section{Data-Augmentation}

In the absence of an adequate volume of training data, it is possible to increase the proper size of existing data through data augmentation, which has contributed to significantly improving the convolutional neural network's performance in the domain of image classification. In the case of ECG signal, data-augmentation (DA) can be applied to signal or spectrograms \citep{Park2019, Pan2020}. In this paper, a DA was applied directly to the spectrogram images to modify and improve the input images.
Spectrogram masking blocks of consecutive frequency channels and masking blocks of utterances were modified. These augmentations have been chosen to help the network be robust against deformations in the time direction, partial loss of frequency information, and partial loss of small ECG segments of the input.

In this paper, it has been applied the data-augmentation based on the imagaug library \citep{imgaug}  to perform the various augmentation operations on images. Their respective values used are Zoom range: 0.15, Width shift: 0.15, Fill mode: nearest, Brightness range: [0.5,1.5], Rotation angle: 10, Height shift: 0.10 and CoarseDropout: True. An important function is the “CoarseDropout” with a probability from 0.2 to 0.5. Drop 2\% of all pixels by converting them to black pixels but do that on a lower-resolution version of the image with 50\% of the original size, leading to $2\times2$ squares being dropped.

It must be mentioned that the augmented train dataset was used only to train the proposed CapsNet model, as it is the main focus of this work, in order to evaluate the performance of the model when it is trained with a bigger and more diverse dataset.

\section{Experiments}

To avoid obtaining biased results, all the results presented in this section were obtained using five-fold (holdout) cross-validation using 80\% of the data for training and 20\% for the test. Moreover, all the results presented in this section were obtained using a CPU Core i7-6700K 4.2 GHz with 32 GB ram and GPU Tesla K20, with Ubuntu 16.04, CUDA 9.0 and TensorFlow r1.8. All the metrics presented in this chapter were obtained using the formulas presented in \citep{Sokolova}.

\subsection{Experiment 1.- The Proposed Capsule Neural Network for Drowsiness Detection}
The proposed CapsNet model for drowsiness detection was used to analyze the data in the 'ULg Multimodality Drowsiness Database' (also called DROZY) \citep{Massoz}. This dataset was obtained by the Laboratory of Exploitation of Signals and Images (INTELSIG).

\begin{figure*}[h!]
\centering
\includegraphics[scale=0.8]{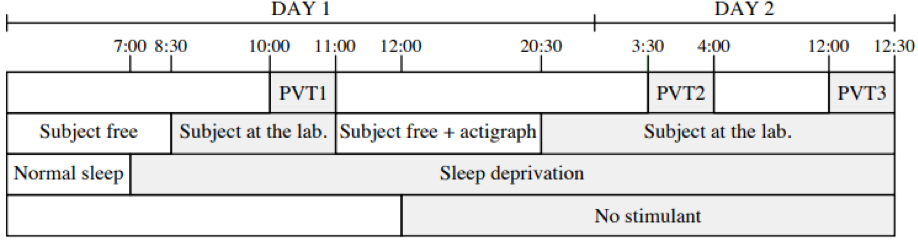}
\caption{Summary of data collection schedule \cite{Massoz}.}
\label{figure7}
\end{figure*}

The database was collected from 14 completely healthy young subjects who were subjected to three psychomotor vigilance tests (PVT1, 2 and 3) of 10 minutes duration, where the subjects had to monitor a red rectangular box over a black background on a computer screen and to press a button as soon as a yellow figure appeared inside the red rectangle. To ensure the drowsy state of the subjects, they were induced into a prolonged waking state during the tests. Figure \ref{figure7} shows the schedule of the tests and also the sleep deprivation times used to carry out the experiment.

For each subject and psychomotor vigilance test, the database contains:
\begin{itemize}
    \item The Karolinska sleepiness scale scores (KSS).
\item	The stimulus and reaction times of the tests.
\item	Five EEG channels (Fz, Cz, Pz, C3, and C4) sampled at 512 Hz. 
\item	Two electrooculography channels (EOG) sampled at 512 Hz. 
\item	Electrocardiogram (ECG) sampled at 512 Hz. 
\item	Electromyography (EMG) sampled at 512 Hz.
\item	Kinect v2 sensor images.
\item	Kinect v2 sensor videos and face signals.

\end{itemize}
The DROZY database is perfectly synchronized over time \cite{Massoz}.

The CapsNet model was able to deal with the spectrogram from EEG with a resolution of $32\times32$ pixels for a single "Fz channel" and also with $64\times32$ pixels when the vertical concatenation of channels Fz (at the top) and Pz (at the bottom) were used. 

A grid search was performed over all the hyperparameters for a Capsule Neural Network (which is described later in this section) in order to analyze different architectures. 

The summary of the proposed models’ layers and architecture is described as follows:

\begin{enumerate}
\item 	One convolutional layer with 64 filters and a kernel size of $5\times5$ with stride 2, which leads to 64 features maps. 

\begin{itemize}
    \item	For $32\times32$ images, each feature map has a size of $14\times14$.
    \item For $64\times32$ images, each feature map has a size of $30\times14$. 
\end{itemize}
\item 	A second convolution layer with 128 filters and a kernel size of $9\times9$ with stride 2 is applied.
\item 	The Primary Capsules layer, consisting of 8 capsules with a dimension of 16. Each resulting Primary Capsules layer has a feature map size of $3\times3$ for input images of $32\times32$ and $11\times3$ for the input images of $64\times32$.
\item 	After the Primary Capsules are formed, each vector is resized using Equation (2), in order to ensure that the length is between 0 and 1.
\item 	The prediction of each capsule in the Primary Capsule layer is obtained applying the dynamic routing process 5 times, which is performed using Equations (3), (4), (5) and (6).
\item 	The prediction was performed by the dynamic routing process that leads into the Secondary Capsules, which consists of 2 capsules, referred to the base and drowsy states, each one with a dimension of 10. In this step, the length of the instantiation vector of each secondary capsule represents the probability of each state. The desired classification is made considering the highest probability value.
\item 	The output of each Secondary Capsule is also fed into a decoder which consists of three fully connected layers:
\begin{itemize}
\item	The first with 512 neurons and ReLu activation.
\item	The second with 1024 and also ReLu activation 
\item	The third fully connected layer has a number of neurons equal to the size of the input data (i.e. either $32\times32$ or $64\times32$) with Sigmoid activation. 
\end{itemize}
\item 	Finally, the squared differences between the output of the decoder and the original input image are calculated, scaled by 0.0005 and added to the margin loss during the training in order to avoid the overfitting of the network. 

All the values used for the hyperparameters were obtained through a grid search. In the first place, the number of feature maps of each convolutional layer was selected testing from 64 up to 256 features maps. The kernel size of the convolutional layers was selected among five values (3, 5, 7, 9 and 11). 
For the number of primary capsules and its dimensions, three different options were considered (8, 16, and 32) according to the number of feature maps of the second convolutional layer. 
Analogously, the dimension of the secondary capsules was tuned by trying out four different values (10, 16, 20 and 24). The number of routing iterations was also tuned considering 5 different options (sequence from 1 up to 5 with a step of 1), which resulted in a value of 5 (in contrast to the 3 routing iterations as used in Sabour et al. \citep{NIPS2017_6975}). Finally, for the reconstruction error, six options were considered (from 0 to 0.00005, dividing by 10 in each step).

\end{enumerate}

\begin{figure*}[h!]
\centering
\includegraphics[scale=0.8]{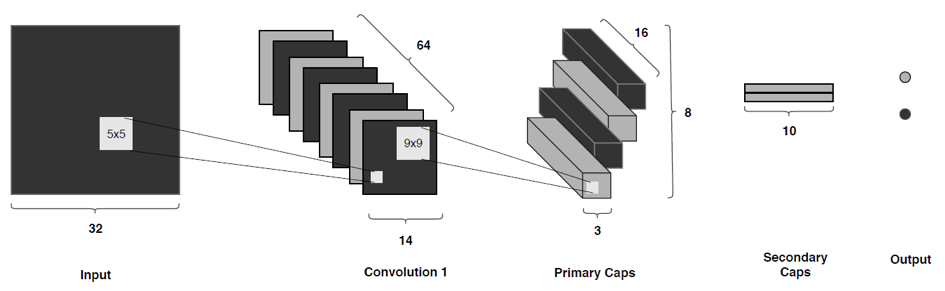}
\caption{CapsNet architecture for 32x32 spectrograms.}
\label{figure6}
\end{figure*}

\subsection{Experiment 2.- Convolutional Neural Network Model}

To assess the effectiveness of the proposed CapsNet model in the drowsiness detection task, its results are compared to the ones obtained from a CNN model. In fact, it is proposed a CNN model with architecture based on Yu-Chia Hung \citep{Hung}. The original network in \citep{Hung}  was developed to face the same classification problem as here, however using a different (non-public) database and with a different input signal shape. Thus, the architecture developed by Yu-Chia Hung was modified to allow input images with a size of $32\times32$, corresponding to the single Fz channel spectrograms, and images of $64\times32$ pixels, corresponding to the vertical concatenation of the spectrograms of the Fz and Pz channels.

\begin{figure*}[h!]
\centering
\includegraphics[scale=0.8]{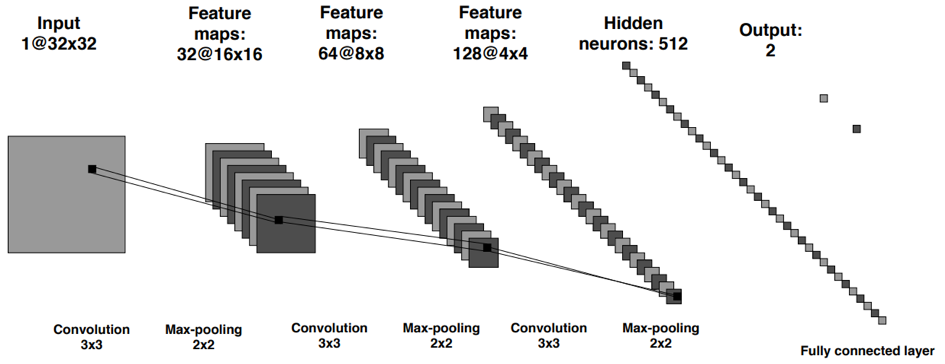}
\caption{CNN architecture for $32\times32$ spectrograms.}
\label{figure10}
\end{figure*}

Indeed, after the input layer, the proposed CNN consists of a $3\times3$ convolutional layer with 32 features maps, followed by a max-pooling layer with a size of $2\times2$ or $4\times2$ depending the case (i.e., $2\times2$ for $32\times32$ images and $4\times2$ for $64\times32$ images) using a stride of 2 and zero-padding. This structure of convolution-pooling layers is replicated three times, maintaining the size for the pooling layers and varying only the number of features maps for the convolutional layers: 32, 64 and 128 for the first, second and third convolutional layers, respectively. Then, the CNN model has a single fully connected layer with 512 hidden neurons, which is responsible for carrying out the desired drowsiness detection. Figure \ref{figure10} shows the CNN’s architecture for dealing with $32\times32$ images and can be summarized as:

\begin{multline}
{Input}=[Z \times 32]-32C[3\times3]-32P[Ax2]-64C[3\times3]\\ -64P[Ax2]-128C[3\times3]-128P[Ax2]-FC[512]-Output[2] 
\end{multline}

where $C$ indicates the convolutional layers; P the pooling layers and FC the fully connected layer. The number before each letter indicates the number of filters of each layer; and the number in brackets shows the size of each filter or the number of hidden neurons. Depending on the data set used as input (i.e., $32\times32$ or $64\times32$), the letter Z has a value of 32 or 64 and, in the same way has a value of 2 or 4.

Also, to avoid the overfitting of the network, Dropout with keep probability of 0.5 is used in the fully connected layer and $L2$ regularisation with parameter $\beta$ of 0.001 was applied to the cost function. These parameters, like the size of the kernel of the convolutional layers and the number of hidden neurons in the classification layer, were obtained through a grid search. For the dropout value, we searched among seven options (sequence from 0 to 0.6 with a step of 0.1), and for the $\beta$ value were explored six options from 0 to 0.0001, dividing by 10 in each step. On the other hand, the size of the kernel of the convolutional layer was picked among three options ($3\times3$, $5\times5$ and $7\times7$), and four options were evaluated for the number of neurons in the classification layers (128, 256, 512 and 1024). Finally, ADAM optimiser was employed, and 500 epochs were used for the model training.

\section{Results and Discussion}

\subsubsection{ CapsNet and CNN models}
Table \ref{table2} shows the average results with their respective standard deviations for each performance metric in the detection of subjects under drowsiness for both datasets Fz and Fz-Pz, and considering the proposed CapsNet model and the CNN model even with Data-Augmentation. 

\begin{table}[]
\centering
\scriptsize
\caption{Summary of performance metrics for the different datasets using CNN and CapsNet. DA represents Data-Augmentation Technique.}
\label{table2}
\resizebox{12cm}{!}{
\begin{tabular}{|c|c|c|c|c|c|c|c|c|c|c|c|c|c|c|c|c|}
\hline
\multirow{2}{*}{\textbf{Dataset}} & \multirow{2}{*}{\textbf{Model}}                        & \multicolumn{3}{c|}{\multirow{2}{*}{\textbf{\begin{tabular}[c]{@{}c@{}}Accuracy\\ (\%)\end{tabular}}}} & \multicolumn{3}{c|}{\multirow{2}{*}{\textbf{\begin{tabular}[c]{@{}c@{}}F1 Score\\ (\%)\end{tabular}}}} & \multicolumn{3}{c|}{\multirow{2}{*}{\textbf{\begin{tabular}[c]{@{}c@{}}Sensitivity\\ (\%)\end{tabular}}}} & \multicolumn{3}{c|}{\multirow{2}{*}{\textbf{\begin{tabular}[c]{@{}c@{}}Specificity\\ (\%)\end{tabular}}}} & \multicolumn{3}{c|}{\multirow{2}{*}{\textbf{\begin{tabular}[c]{@{}c@{}}Precision\\ (\%)\end{tabular}}}} \\
                                  &                                                        & \multicolumn{3}{c|}{}                                                                                  & \multicolumn{3}{c|}{}                                                                                  & \multicolumn{3}{c|}{}                                                                                     & \multicolumn{3}{c|}{}                                                                                     & \multicolumn{3}{c|}{}                                                                                   \\ \hline
\multirow{3}{*}{Fz}               & CNN                                                    & 79,46                             & +/-                             & 4,60                             & 79,55                             & +/-                             & 4,36                             & 80,80                              & +/-                              & 3,95                              & 78,07                              & +/-                              & 5,41                              & 78,30                              & +/-                             & 4,96                             \\ \cline{2-17} 
                                  & \begin{tabular}[c]{@{}c@{}}Caps\\ Net\end{tabular}     & 84,45                             & +/-                             & 0,62                             & 83,76                             & +/-                             & 0,65                             & 85,81                              & +/-                              & 0,97                              & 83,26                              & +/-                              & 0,91                              & 81,81                              & +/-                             & 0,81                             \\ \cline{2-17} 
                                  & \begin{tabular}[c]{@{}c@{}}Caps\\ Net-DA\end{tabular} & \multicolumn{1}{l|}{89,12}        & \multicolumn{1}{l|}{+/-}        & \multicolumn{1}{l|}{0,90}        & \multicolumn{1}{l|}{87,65}        & \multicolumn{1}{l|}{+/-}        & \multicolumn{1}{l|}{0,87}        & \multicolumn{1}{l|}{89,60}         & \multicolumn{1}{l|}{+/-}         & \multicolumn{1}{l|}{0,90}         & \multicolumn{1}{l|}{88,65}         & \multicolumn{1}{l|}{+/-}         & \multicolumn{1}{l|}{0.88}         & \multicolumn{1}{l|}{87,65}         & \multicolumn{1}{l|}{+/-}        & \multicolumn{1}{l|}{0,85}        \\ \hline
\multirow{3}{*}{Fz-Pz}            & CNN                                                    & 75,86                             & +/-                             & 2,39                             & 76,93                             & +/-                             & 3,24                             & 79,47                              & +/-                              & 5,54                              & 71,98                              & +/-                              & 2,82                              & 74,67                              & +/-                             & 2,68                             \\ \cline{2-17} 
                                  & \begin{tabular}[c]{@{}c@{}}Caps\\ Net\end{tabular}     & 86,74                             & +/-                             & 1,57                             & 85,97                             & +/-                             & 1,53                             & 87,57                              & +/-                              & 4,67                              & 86,53                              & +/-                              & 4,29                              & 85,20                              & +/-                             & 3,52                             \\ \cline{2-17} 
                                  & \begin{tabular}[c]{@{}c@{}}Caps\\ Net-DA\end{tabular}  & 91,24                             & +/-                             & 0,95                             & 90,25                             & +/-                             & 0,85                             & 89,65                              & +/-                              & 0,90                              & 86,65                              & +/-                              & 0.91                              & 87,65                              & +/-                             & 0.87                             \\ \hline

\end{tabular}
}
\end{table}

\begin{figure*}[h!]
\centering
\includegraphics[scale=0.8]{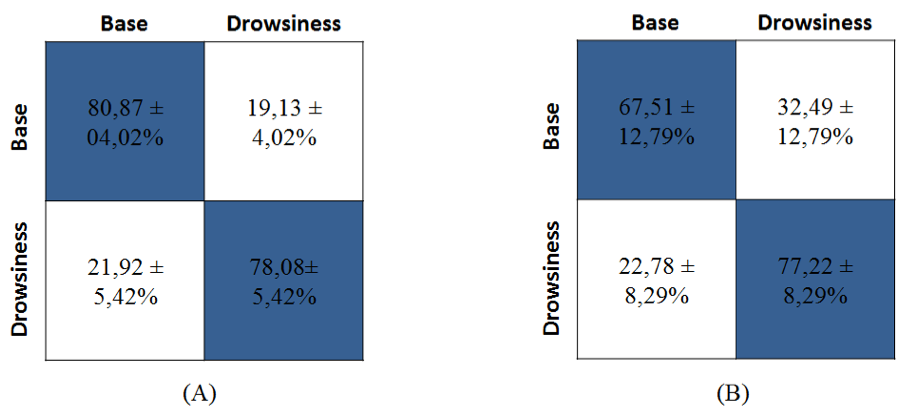}
\caption{Confusion matrices without Data-augmentation for the CNN models using (A) single Fz channel spectrograms and (B) the vertical concatenation of the Fz-Pz channels.}
\label{figure11}
\end{figure*}

\begin{figure*}[h!]
\centering
\includegraphics[scale=0.8]{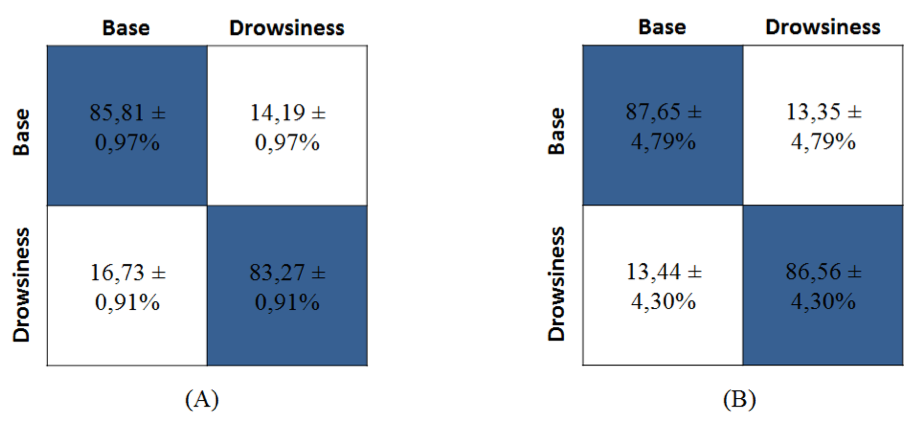}
\caption{Confusion matrices without Data-augmentation for the CapsNets models using (A) single Fz channel spectrograms and (B) the vertical concatenation of the Fz-Pz channels.}
\label{figure12}
\end{figure*}

Figure \ref{figure11} and Figure \ref{figure12} outlines the confusion matrices for both the CNN and CapsNet, respectively, for the two datasets: Fz and Fz-Pz. The values presented in Figure \ref{figure11} and Figure \ref{figure12} are horizontally normalized by class. Note that the results in Table \ref{table2} were obtained from these confusion matrices.

As it can be seen in Table \ref{table2}, the proposed CapsNet model reaches better average results than CNN for both datasets for all the performance metrics with and without data-augmentation. Indeed, CapsNet reaches 84.45\% of accuracy for the Fz dataset and 86.74\% for the dataset Fz-Pz dataset, versus the 79.46\% and 75.86\% reached by the CNN, respectively. Moreover, sensitivity is an important metric in drowsiness detection since it represents the ability of a model to detect subjects who are effectively in a state of drowsiness. 

The CapsNet has a sensitivity score of 85.81\% for the Fz dataset and 87.57\% for the Fz-Pz dataset versus the 80.80\% and the 79.47\% obtained by the CNN, respectively. These results are also corroborated in Figure \ref{figure11} and in Figure \ref{figure12}, where the confusion matrices show that the CapsNet is better than CNN in detecting drowsy subjects who are really with drowsiness.

The proposed CapsNet model also gives smaller standard deviations for all the performance metrics. For instance, CapsNet model has a standard deviation of 0.97\% for sensitivity, which is the maximum standard deviation value for all the evaluated metrics in the Fz dataset. On the other hand, the standard deviation for the sensitivity in case of the CNN model in the Fz dataset is 3.95\%, which is more than four times greater than the one obtained by the CapsNet. These results corroborate the observation that the CapsNet model has a better performance when dealing with smaller datasets, which is an important advantage when working with biomedical signals.

Furthermore, it can be observed that there are differences in the results from each model using the different datasets. Indeed, the CapsNet model gives better average results for each of the evaluated performance metrics when processing the dataset with the joint use of the Fz-Pz channels. However, this increase in the average comes at a cost of an increase in the standard deviation of each one of the metrics. Also, note that the CNN model’s performance deteriorates in terms of all metrics in Table 2 when detecting drowsy individuals based on the fusion of Fz and Pz electrodes.

It should be mentioned that the use of only the Pz channel was not considered since in our experiments, its isolated use only reached a 64.78\% of accuracy, which would not allow the detection of subjects with drowsiness, which is consistent with previous research \citep{JACOBEDENAUROIS201995}. Indeed, the different results for each dataset can be explained by two main facts. Firstly, as demonstrated in \citep{Leonard_Trejo}, the presence of drowsiness in a subject is related to an increase of power in theta waves (tracked by the Fz channel) and a decrease in alpha power (tracked by the Pz channel). Thus, when a subject is under drowsiness, he or she suffers a large phasic increase (event-related) in theta waves, but a tonic decrease (related to life cycle or psychological state) in alpha waves. Moreover, the alpha frequency band has large inter-individual differences associated to age and memory performance \citep{KLIMESCH1999169}.

Based on these results, one can argue that although the Pz channel provides useful information regarding the state of drowsiness of the subject, this information is related to inherent characteristics of each subject, (as age, gender or mental state). Thus, the model cannot effectively use the signal from the Pz channel to generalise between different subjects, and the drowsiness detection model interprets some characteristics delivered by the Pz channel as noise, leading to a less stable model.

\subsubsection{Comparison to shallow models}
To have a fair basis of comparison, the SVM and the Random Forest models were submitted to hyperparameter optimisation of the SVM and the Random Forest by means of grid search. As a result, a linear kernel and a penalisation cost of 0.1 were obtained for the SVM. The Random Forest model was trained using 1,024 trees, a max depth of 25 and a min sample leaf of 1. To assess and quantify the importance and quality of automatic features extraction done by the Deep Learning models (CapsNet and CNN) from EEG signals, the shallow Neural Network model in this section was trained using only one hidden layer of 512 neurons.

Table \ref{table3} shows the performance of three shallow machine learnings models (Support Vector Machines, Random Forest and a Neural Network) for the same classification task based on the same datasets. For them, the input is the individual pixels of the corresponding spectrograms.

\begin{table*}[]
\centering
\caption{Summary of performance metrics for the different datasets using SVM, RF, and NN.}
\label{table3}
\resizebox{\textwidth}{!}{%
\begin{tabular}{|c|c|c|c|c|c|c|c|c|c|c|c|c|c|c|c|c|}
\hline
\multirow{2}{*}{\textbf{Dataset}} & \multirow{2}{*}{\textbf{Model}} & \multicolumn{3}{c|}{\multirow{2}{*}{\textbf{\begin{tabular}[c]{@{}c@{}}Accuracy\\ (\%)\end{tabular}}}} & \multicolumn{3}{c|}{\multirow{2}{*}{\textbf{\begin{tabular}[c]{@{}c@{}}F1 Score\\ (\%)\end{tabular}}}} & \multicolumn{3}{c|}{\multirow{2}{*}{\textbf{\begin{tabular}[c]{@{}c@{}}Sensitivity\\ (\%)\end{tabular}}}} & \multicolumn{3}{c|}{\multirow{2}{*}{\textbf{\begin{tabular}[c]{@{}c@{}}Specificity\\ (\%)\end{tabular}}}} & \multicolumn{3}{c|}{\multirow{2}{*}{\textbf{\begin{tabular}[c]{@{}c@{}}Precision\\ (\%)\end{tabular}}}} \\
                                  &                                 & \multicolumn{3}{c|}{}                                                                                  & \multicolumn{3}{c|}{}                                                                                  & \multicolumn{3}{c|}{}                                                                                     & \multicolumn{3}{c|}{}                                                                                     & \multicolumn{3}{c|}{}                                                                                   \\ \hline
\multirow{3}{*}{Fz}               & SVM                             & 67.28                             & +/-                             & 3.06                             & 66.13                             & +/-                             & 3.65                             & 63.85                              & +/-                              & 4.13                              & 70.84                              & +/-                              & 6.00                              & 69.94                              & +/-                             & 6.07                             \\ \cline{2-17} 
                                  & RF                              & 73.58                             & +/-                             & 3.39                             & 72.00                             & +/-                             & 3.66                             & 68.12                              & +/-                              & 6.42                              & 79.71                              & +/-                              & 8.81                              & 77.48                              & +/-                             & 8.62                             \\ \cline{2-17} 
                                  & NN                              & \multicolumn{1}{l|}{69.59}        & \multicolumn{1}{l|}{+/-}        & \multicolumn{1}{l|}{2.46}        & \multicolumn{1}{l|}{68.11}        & \multicolumn{1}{l|}{+/-}        & \multicolumn{1}{l|}{4.11}        & \multicolumn{1}{l|}{65.62}         & \multicolumn{1}{l|}{+/-}         & \multicolumn{1}{l|}{7.73}         & \multicolumn{1}{l|}{72.97}         & \multicolumn{1}{l|}{+/-}         & \multicolumn{1}{l|}{7.10}         & \multicolumn{1}{l|}{71.29}         & \multicolumn{1}{l|}{+/-}        & \multicolumn{1}{l|}{2.10}        \\ \hline
\multirow{3}{*}{Fz-Pz}            & SVM                             & 71.30                             & +/-                             & 3.67                             & 72.59                             & +/-                             & 1.83                             & 72.16                              & +/-                              & 3.46                              & 69.58                              & +/-                              & 9.88                              & 73.25                              & +/-                             & 3.86                             \\ \cline{2-17} 
                                  & RF                              & 72.50                             & +/-                             & 4.90                             & 71.00                             & +/-                             & 5.63                             & 68.28                              & +/-                              & 7.58                              & 76.68                              & +/-                              & 2.82                              & 74.11                              & +/-                             & 3.49                             \\ \cline{2-17} 
                                  & NN                              & 67.28                             & +/-                             & 3.06                             & 66.13                             & +/-                             & 3.65                             & 63.85                              & +/-                              & 4.13                              & 80.69                              & +/-                              & 7.31                              & 74.99                              & +/-                             & 7.26                             \\ \hline
\end{tabular}%
}
\end{table*}

Based on the results reported in Table \ref{table3}, none of these models reaches an accuracy higher than 80\% in the drowsiness detection. In contrast, the proposed CapsNet model delivered an accuracy of 86.74\%. Moreover, the shallow models have, on average, standard deviations that are significantly greater than both CapsNet and CNN models. Indeed, the higher standard deviation value for each shallow model in the Fz dataset (considering all the evaluated metrics) was 6.07\%, 8.81\% and 7.73\% for the SVM, Random Forest and Neural Network, respectively. If we compare these values with the standard deviation obtained by the CapsNet model, we have between six and nine times greater standard deviation for the shallow models than for the CapsNet model: the shallow models have a higher variability and a diminished capability of generalization.

Based on these results, it can be inferred that the automatic feature extraction made by the convolution layers is essential when working with spectrograms images from EEG signals. In particular, the shallow neural network trained with the same architecture for the hidden layer as in the CNN model presents a significantly variable performance, which corroborates the importance of the use of convolution layers for the feature extraction. In this sense, the CapsNet model takes advantage of the convolutional layer to extract representative features from the EEG spectrograms and also can handle small datasets through the dynamic routing algorithm, delivering a smaller model variability.

Furthermore, the results in Table \ref{table3} present a variability of up to 10\% between accuracy and precision, which indicates that the shallow models cannot handle the drowsiness detection in a balanced way. That is, despite that the shallow models can reach a specificity of 79.71\%, as it is the case of the Random Forest model for the Fz dataset, the same model reaches a sensitivity of only 68.12\% for the same dataset, hence, it would be able to correctly classify the subjects without drowsiness, but it would not be able to detect subjects with drowsiness.

Table 4, present a comparison among drowsiness detection models using "ULg Datasets".
Based on Table \ref{tab:comparison}, one can observe that the proposed model achieves higher accuracy than all the other models except by the model proposed by Garcia et.al \citep{garcia} which uses video as input data. In this sense, we can divide the models from Table IV in three groups: the ones that need a camera to obtain the input data (image or video), the models that uses biological data as input, and finally the model from by Massoz et. al \citep{Maftukhaturrizqoh_2019}, which originally presented the “ULg Multimodality Drowsiness Database” and that uses the time reaction from the psychomotor vigilance tests to detect drowsiness. 
First, one should note that the model presented by Massoz et. al \citep{Maftukhaturrizqoh_2019} can just be used as a comparison, because the time of reactions is not a data that can be collected continuously in a real car driving situation or in a work environment, so it is not applicable for real-time detection. On the other hand, the model proposed by Maftukhaturrizqoh et.al \citep{Maftukhaturrizqoh_2019} uses electrocardiography (ECG) signals to achieve de detection of drowsiness, using various handcrafted features from the ECG signals, such as the time interval between two consecutive heartbeats. Although this model takes into account the operator privacy (not using any camera device), its accuracy is below the values reached by our proposed model.
Finally, from the two models proposed by Garcia et.al \citep{garcia}, the one that uses facial images as input reached a lower accuracy than our model, but the one that uses video sequences achieve higher results. However, these two models have two main limitations. In the first place, both models require a video device in front of the subject, which makes these techniques too invasive to the subject’s privacy. In second place, an investigation conducted by Ngxande et.al \citep{Ngxande} shows that models that are trained for drowsiness detection with public databases such as the “ULg Multimodality Drowsiness Database” and use image or video sequences as input, do not generalize well when are tested on dark-skinned people due to the bias that the databases can have in terms of ethnical diversity. In this context, in the work conducted by Garcia et.al \citep{garcia}, when they tested their model with real-world data, they achieved a much lower accuracy of 68.7\% using images as input and a 72.73\% of accuracy using video sequences, which corroborates the work presented in Ngxande et.al \citep{Ngxande}. The sixth row of Table 4 shows our best results with and without data-augmentation.

\begin{table}[H]
\centering
\scriptsize
\caption{Comparison among drowsiness detection models using the “ULg Multimodality Drowsiness Database”. DA represents Data Aumentation.}
\label{tab:comparison}
\begin{tabular}{|c|c|c|c|}
\hline
\textbf{Author}                                                                                                   & \textbf{Metric}                                           & \textbf{Classifier}                                                      & \textbf{\begin{tabular}[c]{@{}c@{}}Accuracy\\ (\%)\end{tabular}} \\ \hline
\begin{tabular}[c]{@{}c@{}} \citep{Ngxande}\end{tabular}                            & \begin{tabular}[c]{@{}c@{}}Time\\ Reaction\end{tabular}   & \begin{tabular}[c]{@{}c@{}}Suppor\\ Vector\\ Machine\end{tabular}        & 78.46                                                            \\ \hline
\begin{tabular}[c]{@{}c@{}} \citep{Maftukhaturrizqoh_2019}\end{tabular} & \begin{tabular}[c]{@{}c@{}}ECG\\ Features\end{tabular}    & \begin{tabular}[c]{@{}c@{}}Neural\\ Network\end{tabular}                 & 81.60                                                            \\ \hline
\begin{tabular}[c]{@{}c@{}} \citep{garcia}\end{tabular}                             & \begin{tabular}[c]{@{}c@{}}Subject\\ Image\end{tabular}   & \begin{tabular}[c]{@{}c@{}}Convolutional\\ Neural\\ Network\end{tabular} & 85.68                                                            \\ \hline
\begin{tabular}[c]{@{}c@{}} \cite{garcia}\end{tabular}                             & \begin{tabular}[c]{@{}c@{}}Subject\\ Video\end{tabular}   & \begin{tabular}[c]{@{}c@{}}Convolutional\\ Neural\\ Network\end{tabular} & 91.07                                                            \\ \hline
\textbf{Proposed Model}                                                                                           & \begin{tabular}[c]{@{}c@{}}EEG\\ Spectrogram\end{tabular} & \begin{tabular}[c]{@{}c@{}}Capsule\\ Network\end{tabular}                & 86.74 / 91.24 (DA)                                                           \\ \hline
\end{tabular}
\end{table}

Overall, it can be argued that, of all the above deep and shallow models, the CapsNet model is the only one capable of a stable drowsiness detection as it has a standard deviation smaller than 1\%. The relevance of this result lies in the amount of data available to train the models (920 samples in total), as it is usually the challenging case when dealing with biomedical data, which in turn represents a severe limitation when developing Deep Learning models.

\section{Conclusions} 

The detection of drowsiness is currently one of the most relevant challenges to guarantee safety and reliability in several industry sectors. 

Another challenge involving drowsiness detection is the paucity of biomedical data that imposes severe limitations when developing deep learning models for drowsiness detection. These models require massive amounts of data for the learning and training process.

As a tentative to circumvent this challenge, this work presented a new approach using the Capsule Neural Network model to perform drowsiness detection. Based on the results, with and without data-augmentation presented in Section 4, the proposed CapsNet model is capable of delivering better results than CNN and other shallow machine learning models such as Support Vector Machines, Random Forest and Neural Networks to work with small datasets and transient biological signals such as EEG. 
This is partially due to the capsules’ ability and advantages in working with small amounts of data and because they do not have the positional invariance problems of CNNs as well as the capability of CapsNet to keep the spatial and hierarchical relationships among the extracted features, which results in improved generalization capacity.

Also, it was shown that spectrograms from the frontal (Fz) and parietal (Pz) electrodes could be used as inputs to the drowsiness detection models due to the relationship between the level of drowsiness of a subject and the variations in the amplitude in the range of theta and alpha waves tracked by the frontal (Fz) and parietal (Pz) channels, respectively. In this context, despite the combined use of the frontal (Fz) and parietal (Pz) signals’ spectrograms achieved slightly superior results in all the evaluated metrics when compared to the spectrograms obtained from the signals tracked by the frontal (Fz) electrode alone, the use the single frontal channel provides standard deviations up to four times lower in the evaluated metrics than the results from the fusion of both electrodes, thus suggesting that it represents a more stable and reliable model for drowsiness detection. 

On the other hand, since the most challenging step in the development of Deep Learning applications is the training process, which requires high computational resources (characteristics of the used machine for training are described in Section 4), after this step, the already trained model can be easily imported into a portable computer or mobile phone, since the trained model does not require any specific overwhelming requirements to operate, allowing the implementation of a wearable drowsiness identification system.

Therefore, based on the results reported in this study, it can be argued that Capsule Neural Networks are a promising approach for drowsiness detection and are worth further research and development.

\section{Future Work} As future work, a new database will be developed in order to study the influence of Occipital electrodes' signals and how this information allows us to improve the drowsiness detection.

\bibliography{references}
\vspace{-1.5cm}
\end{document}